\documentclass[lettersize,journal]{IEEEtran}
\usepackage{amsmath,amsfonts}
\usepackage{algorithmic}
\usepackage{algorithm}
\usepackage{array}
\usepackage[caption=false,font=normalsize,labelfont=sf,textfont=sf]{subfig}
\usepackage{textcomp}
\usepackage{stfloats}
\usepackage{url}
\usepackage{verbatim}
\usepackage{graphicx}
\usepackage{cite}
\usepackage{multirow} 
\usepackage{pifont}
\usepackage{multirow}
\usepackage[table,xcdraw]{xcolor}
\usepackage{colortbl} 
\usepackage{amsmath, amssymb} % 
\usepackage{graphicx}        
\usepackage{caption}         
\usepackage{booktabs}
\usepackage[table]{xcolor}

\begin{document}
\title{Lightweight Facial Landmark Detection in Thermal Images via Multi-Level Cross-Modal Knowledge Transfer}
\author{Qiyi Tong, Olivia Nocentini, Marta Lagomarsino, Kuanqi Cai, Marta Lorenzini, and Arash Ajoudani

\thanks{The authors are with the Human-Robot Interfaces and Interaction Laboratory, Istituto Italiano di Tecnologia, Genoa, Italy. 
Qiyi Tong is also with the Ph.D. Program of National Interest in Robotics and Intelligent Machines (DRIM), Università di Genova, Genoa, Italy.}
\thanks{Corresponding author's email: {\tt\small qiyi.tong@iit.it}}
}

% \thanks{Corresponding author's email: {\tt\small qiyi.tong@iit.it}}
% \thanks{The authors are with the Human-Robot Interfaces and Interaction Laboratory, Istituto Italiano di Tecnologia, Genoa, Italy.
% Qiyi Tong is also with the Ph.D. Program of National Interest in Robotics and Intelligent Machines (DRIM), Università di Genova, Genoa, Italy.}

% \title{MLCM-KD: Multi-Level Cross-Modal Knowledge Distillation from RGB to Thermal for Facial Landmark Detection}

% \title{Dual-Injected Cross-Modal Distillation: Multi-Level Knowledge Transfer for Lightweight Thermal Facial Landmark Detection}



\maketitle

\begin{abstract}
Facial Landmark Detection (FLD) in thermal imagery is critical for applications in challenging lighting conditions, but it is hampered by the lack of rich visual cues.
Conventional cross-modal solutions, like feature fusion or image translation from RGB data, are often computationally expensive or introduce structural artifacts, limiting their practical deployment. To address this, we propose Multi-Level Cross-Modal Knowledge Distillation (MLCM-KD), a novel framework that decouples high-fidelity RGB-to-thermal knowledge transfer from model compression to create both accurate and efficient thermal FLD models.
A central challenge during knowledge transfer is the profound modality gap between RGB and thermal data, where traditional unidirectional distillation fails to enforce semantic consistency across disparate feature spaces. To overcome this, we introduce Dual-Injected Knowledge Distillation (DIKD), a bidirectional mechanism designed specifically for this task. DIKD establishes a connection between modalities: it not only guides the thermal student with rich RGB features but also validates the student's learned representations by feeding them back into the frozen teacher's prediction head. This closed-loop supervision forces the student to learn modality-invariant features that are semantically aligned with the teacher, ensuring a robust and profound knowledge transfer.
Experiments show that our approach sets a new state-of-the-art on public thermal FLD benchmarks, notably outperforming previous methods while drastically reducing computational overhead.

\end{abstract}

\begin{IEEEkeywords}
Facial Landmark Detection, Cross-Modal Knowledge Distillation,  Thermal Images, Computer Vision.
\end{IEEEkeywords}

\section{Introduction}

Facial landmark detection (FLD), the task of precisely locating semantically significant points on a human face (e.g., eye corners, nose tip, mouth contours)\cite{8094282, zhang2014facial, 8910587}, serves as a foundational component for numerous computer vision applications. These include face recognition \cite{juhong2017face, wu2019facial, khabarlak2021fast}, head pose estimation\cite{wang2020fast, zhu2016face}, gaze estimation\cite{yu2018deep, datta2021eye},  alertness detection\cite{jeong2017driver, kim2023real, nocentini2025graph}, and anomaly detection\cite{hariprasad2023boundary, merlo2023automatic}.

In the visible-light spectrum, FLD has achieved remarkable progress on RGB images, particularly with the development of increasingly sophisticated neural architectures. Early regression-based methods like DeepPose \cite{toshev2014deeppose} and Wing Loss \cite{feng2018wing} directly predict landmark coordinates with compact models and fast inference. However, due to the high degree of freedom in facial poses, these approaches often struggle to capture intricate spatial relationships, limiting localization precision \cite{dubey2023comprehensive}.
To address this, heatmap-based methods have become dominant, reformulating the task as a per-pixel localization problem. By predicting spatial confidence distributions, these methods inherently preserve local structure and demonstrate superior robustness. 
Pioneering works include Convolutional Pose Machines (CPMs) \cite{wei2016convolutional}, which aggregate multi-stage spatial context, and stacked Hourglass networks \cite{newell2016stacked}, which fuse multi-scale information via upsampling and downsampling pipelines. HRNet \cite{sun2019deep} further advances performance by maintaining high-resolution features throughout the network and enabling cross-scale fusion. More recently, methods like SimCC \cite{li2022simcc} have recast keypoint localization as a classification task over discretized coordinates, inspiring a new generation of lightweight yet powerful models such as RTMPose \cite{jiang2023rtmpose} and DWPose \cite{yang2023effective} that excel in real-time applications. 
 %Other works tackle occlusion by modeling uncertainty (e.g., ProbPose \cite{purkrabek2025probpose}) or leveraging generative priors via vector quantized autoencoder (VQ-VAE)\cite{van2017neural} (e.g., PCT \cite{geng2023human}).

However, RGB-based methods suffer significant performance degradation in low-light or fully dark environments. Such scenarios are increasingly prevalent in applications such as surveillance, autonomous driving, and security. 
To address this limitation, infrared imaging has emerged as a promising alternative due to its ability to operate independently of ambient lighting conditions \cite{reza2013infrared}. %, operating independently of ambient lighting by capturing emitted long-wave infrared radiation rather than reflected visible light. This grants thermal sensors inherent robustness in challenging lighting conditions. 
Among infrared modalities, there are no long-exposure, validated studies confirming that emissions from active infrared sensors are harmless for prolonged facial exposure. Their deployment in workplace settings could therefore pose potential risks, motivating the investigation of passive thermal sensors as a safer alternative.
%
%Despite this advantage, achieving accurate facial landmark detection in thermal images remains a significant challenge. 
Thermal sensors provide only surface temperature distributions, lacking the texture, color, and high-frequency structural details characteristic of RGB data. As a result, the discriminability between facial keypoints is substantially reduced. %For instance, the small temperature difference (typically 1–2°C) between the canthus and eyelid often manifests as a homogeneous thermal region, making it difficult to delineate precise boundaries\cite{de2023infrared,budzan2013face}. This ambiguity is further exacerbated in low-resolution scenarios, where spatial detail is inherently limited. Consequently, models trained solely on RGB images fail to generalize effectively to the thermal domain, due to the pronounced modality gap \cite{sun2019deep}.

To bridge this gap, prior work has explored image-to-image translation and RGB-thermal fusion. The former converts thermal images into pseudo-RGB representations to exploit models pre-trained on large RGB datasets \cite{brenner2023rgb, bourlai2023data}. While effective in some contexts, this approach is unsuitable for a geometrically precise task like FLD, as translation-induced artifacts, structural distortions, and spatial misalignments are inevitably propagated to landmark prediction, causing systematic localization errors. In contrast, RGB-thermal fusion combines features from both modalities, but it fails under near-total darkness, where the RGB input is dominated by noise, corrupting the feature space and impairing the fused model's performance. Moreover, processing dual streams incurs high computational cost, which conflicts with the increasing demand for lightweight models suitable for resource-constrained edge devices. 
This highlights the need for a method that transfers the rich geometric and semantic knowledge from RGB data to a thermal-only model, while eliminating the dependency on the RGB stream at inference. 

Cross-modal knowledge distillation (KD) emerges as an ideal paradigm. Transferring high-level representations from a powerful RGB teacher to a lightweight thermal student can significantly boost thermal-only performance while maintaining computational efficiency.
However, effective cross-modal KD is challenging. Because RGB captures reflected light whereas thermal imaging measures emitted heat, the two modalities yield inherently different feature spaces with no simple one-to-one correspondence. Naïve distillation strategies that simply match intermediate features or output logits often fail, as they overlook this semantic chasm and can lead to counterproductive knowledge transfer. An effective RGB-to-thermal distillation framework must therefore move beyond simplistic, unidirectional knowledge injection. It demands a mechanism that actively facilitates semantic alignment and accounts for the representational disparities between modalities.

This paper introduces Dual-Injected Knowledge Distillation (DIKD), a novel bidirectional mechanism that establishes structural and semantic alignment between an RGB teacher and a thermal student. By creating a closed-loop information flow where features are injected in both forward (teacher-to-student) and reverse (student-to-teacher) directions, DIKD forces the student to learn feature representations that are not only guided by the teacher but are also interpretable by the teacher. This bidirectional constraint stabilizes training and ensures a more profound and robust knowledge transfer. Furthermore, to meet the requirements of edge deployment, we introduce a second-level model compression phase, which distills the learned knowledge from the trained thermal model into a lightweight version, achieving a favorable trade-off between accuracy and computational efficiency.

In summary, our contributions are threefold:
\begin{itemize}
\item {We propose MLCM-KD, a novel multi-level cross-modal knowledge distillation framework specifically designed for lightweight thermal facial landmark detection.} Unlike previous works that either rely on direct modality fusion or unidirectional distillation, our framework decouples cross-modal knowledge transfer and model compression into two hierarchical stages. This design enables the effective transfer of rich semantic and geometric priors from RGB to thermal, followed by the distillation of a compact, high-performance thermal model suitable for real-world edge deployment.
\item {We introduce DIKD, a bidirectional cross-modal supervision mechanism that explicitly bridges the modality gap between RGB and thermal domains.} By structurally injecting features between the teacher and student prediction heads in both forward and reverse directions, DIKD not only enhances semantic alignment and representation compatibility, but also stabilizes the distillation process. This dual-injection strategy enables the thermal student to both absorb and be interpretable by the RGB teacher, leading to more robust and accurate landmark localization under challenging thermal conditions.
\item {We provide the first comprehensive study of cross-modal distillation for thermal facial landmark detection, establishing new state-of-the-art results on multiple public benchmarks.} Our extensive experiments and ablation studies reveal several novel insights: (1) bidirectional feature-head interaction is critical for effective cross-modal transfer; (2) our approach significantly outperforms both traditional unidirectional distillation and image translation baselines in terms of accuracy and efficiency; and (3) the resulting lightweight models maintain high performance even under severe low-light and resource-constrained scenarios, demonstrating strong practical value for real-world applications.
\end{itemize}

The remainder of this paper is organized as follows: Section \ref{sec:method} presents a detailed description of the proposed MLCM-KD framework. Section~\ref{sec:exp} describes the experimental setup and evaluation protocols. Section~\ref{sec:results} reports and analyzes the experimental results, including benchmark comparisons and ablation studies. Finally, Section \ref{sec:conclusion} concludes the paper with a summary and directions for future work.

\section{METHODOLOGY}
\label{sec:method}

\begin{figure*}[ht]
\centering
\includegraphics[width=\textwidth]{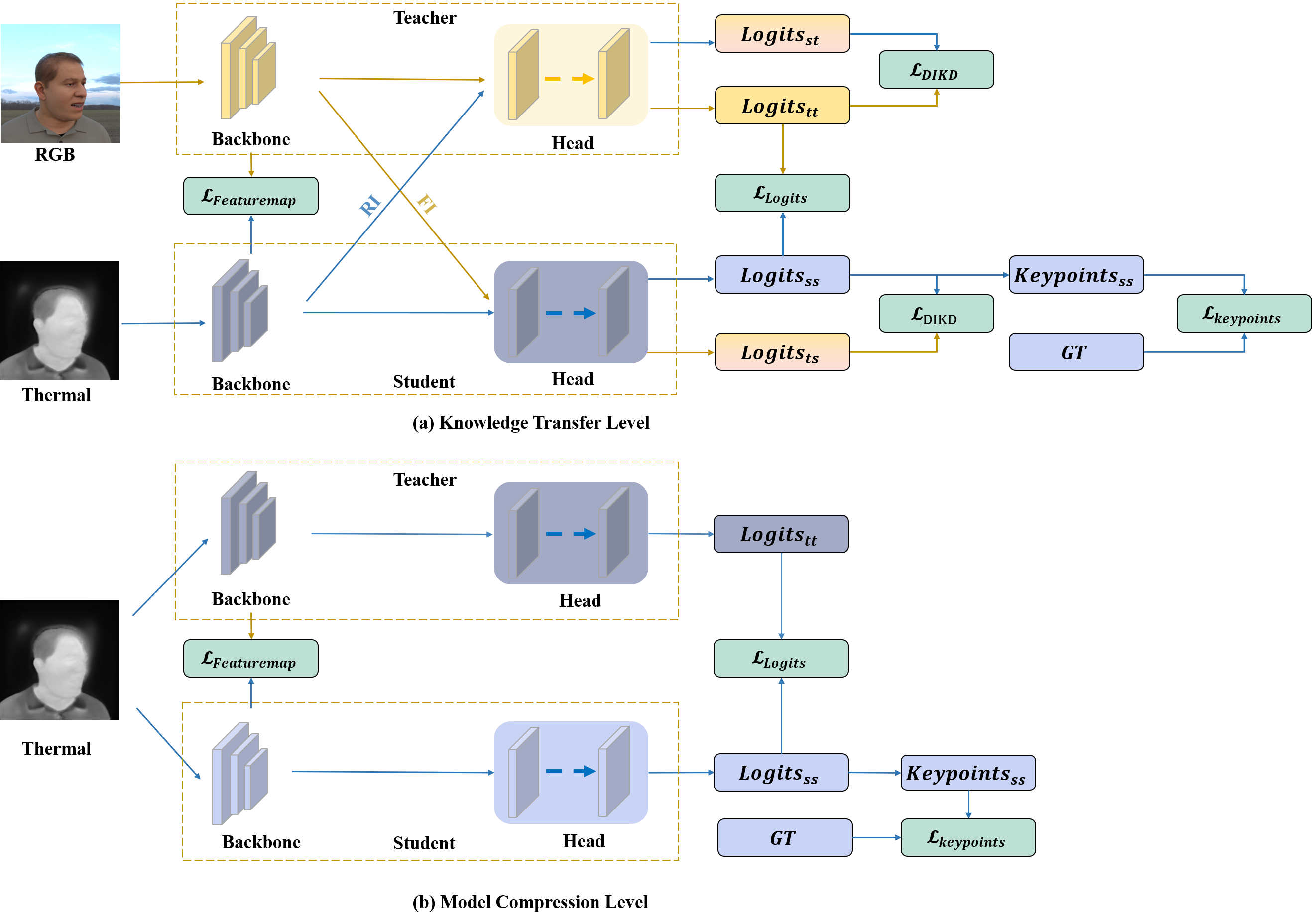}
\caption{Pipeline of Multi-Level Cross-Modal Knowledge Distillation (MLCM-KD), consisting of (a) the Knowledge Transfer Level, which transfers knowledge from RGB to thermal models, and (b) the Model Compression Level, which reduces model complexity for efficient deployment.}
\label{pipeline_KD}
\end{figure*}

In this section, we provide a detailed description of our proposed Multi-Level Cross-Modal Knowledge Distillation (MLCM-KD) framework for lightweight thermal facial landmark detection. As illustrated in Fig.~\ref{exp_KD}, MLCM-KD is composed of two hierarchical levels, each designed to address a specific objective while operating on different input modalities. The first level, named \textit{Knowledge Transfer Level} (KTL), focuses on cross-modal knowledge transfer from RGB to thermal, while the second level, named \textit{Model Compression Level} (MCL), emphasizes model compression using thermal-only inputs for efficient deployment.

\subsection{The Knowledge Transfer Level}
The first level of the MLCM-KD framework named Knowledge Transfer Level (KTL) begins with paired RGB–thermal images as input. At the KTL, landmark-specific knowledge is transferred from a pretrained RGB teacher network to a thermal student network, while actively filtering out modality-sensitive representations to focus on geometric, task-relevant features.
To achieve this comprehensively, the KTL integrates a number of distillation losses operating at different semantic levels: feature mimicry, logit distribution matching, and direct keypoint supervision. Crucially, to address the shortcomings of simple one-way distillation, we introduce our novel Dual-Injected Knowledge Distillation (DIKD) mechanism to enforce a deeper, bidirectional semantic alignment.

% To mitigate the RGB-thermal domain gap and facilitate cross-modal knowledge transfer, we propose the DIKD mechanism. DIKD constructs dual cross-modal pathways: (1) RGB backbone features are injected into the thermal student’s prediction head, and (2) thermal backbone features are fed into the RGB teacher’s prediction head. While primary knowledge flow remains from RGB to thermal, the teacher-head injection enables the RGB network to adapt its guidance to the student’s feature space, stabilizing distillation.
% By injecting intermediate features, DIKD tightens interactions between feature representations and prediction heads. This enhances the thermal student’s spatial-semantic learning for accurate landmark localization, enabling it to distill robust, modality-invariant representations from thermal’s limited information, leveraging RGB’s richer guidance.

\subsubsection{Feature-Based Distillation}
At the intermediate representation level, we encourage the thermal student network to mimic the structural features learned by the RGB teacher. Specifically, given a pair of RGB and thermal images $<I_{RGB}, I_{Thermal}>$, we extract feature maps from the backbone encoders of both teacher and student networks and minimize their discrepancy via an MSE (mean Squared Error) loss:
\begin{equation}
\label{equation_heatmap}
L_{fm} = \frac{1}{CHW} \sum_{c=1}^{C} \sum_{h=1}^{H} \sum_{w=1}^{W} \left( F_{RGB}^{c,h,w} - f\left(F_{Thermal}^{c,h,w}\right) \right)^2 ,
\end{equation}
where $F_{RGB}$ and $F_{Thermal}$ denote the feature maps from the teacher and student backbones, respectively. $C$, $H$, $W$ denote the channel, height and width of the teacher's feature. $f$ represents a $1\times 1$ convolutional layer to reshape the $H_{Thermal}$ to the same dimension as $H_{RGB}$.
This promotes the learning of modality-invariant feature representations and establishes a solid foundation for consistent alignment in predictions.

\subsubsection{Logits-Based Distillation}
The RTMPose series of models, including RTMPose~\cite{jiang2023rtmpose}, RTMW~\cite{jiang2024rtmw}, and RTMO~\cite{lu2024rtmo}, adopt the SimCC-based formulation~\cite{li2022simcc} for keypoint detection, which reformulates the keypoints detection task as a classification problem over discretized coordinate bins along the horizontal and vertical axes.
To enable effective knowledge transfer of spatially relevant cues from the RGB teacher to the thermal student, particularly in the context of facial landmark localization, we adopt a logits-based distillation ($L_{lg}$) loss similar to that employed in DWPose~\cite{yang2023effective}. The loss is defined as:
\begin{equation}
\label{equation_logits_loss}
L_{lg} = -\frac{1}{N}\sum_{n}^{N}\sum_{k}^{K}\sum_{i}^{L} L_{RGB}^i \log (L_{Thermal}^i) ,
\end{equation}
where $N$ is the number of the person samples in a batch. $K$ denotes the number of facial landmarks, which is fixed at 68 in accordance with the annotation standards used in both the 300W~\cite{sagonas2016300} dataset and the facial keypoint definitions in the COCO-WholeBody~\cite{jin2020whole} dataset. $L$ is the number of discrete bins along each coordinate axis (either $x$ or $y$). $L^i_{RGB}$ and $L_{Thermal}^{i}$ represent the classification logits over spatial bins predicted by the teacher and student networks, respectively.
This formulation encourages the student network to learn the spatial confidence distribution of the teacher, enhancing geometric coherence in thermal landmark predictions.

\subsubsection{Keypoints-Based Distillation}
In addition to mimicking the teacher's distribution, we directly supervise the student with the ground-truth annotations from the thermal images to sharpen its localization accuracy. This is a standard supervised loss on the student's predictions:
\begin{equation}
\label{equation_keypoints}
    L_{kp} = - \sum_{n=1}^{N}\sum_{k=1}^{K} W_{n,k} \sum_{i=1}^{L}\frac{1}{L} P^i_{GT} \log (P_{Thermal}^i),
\end{equation}
where $N$, $K$ and $L$ retain the same definitions as in Equation ~\ref{equation_logits_loss}. $W_{n,k}$ is a visibility-aware weight mask used to ignore invisible keypoints during training. $P^i_{GT}$ denotes the one-hot encoded ground-truth distribution over coordinate bins, while $P^i_{Thermal}$ represents the student network’s predicted probability over those bins.
This formulation directly enforces accurate keypoint localization by penalizing spatial misalignment between predicted and annotated landmarks, complementing the logits-level supervision and further enhancing geometric precision in thermal imagery.

\subsubsection{Dual-Injected Based Distillation}
To address the inherent modality discrepancy of RGB and thermal, we propose a novel Dual-Injected Knowledge Distillation (DIKD) mechanism that establishes bidirectional structural alignment between the RGB teacher and thermal student networks. Unlike conventional unidirectional distillation that solely injects knowledge from teacher to student, DIKD introduces two complementary interaction pathways:
% \begin{itemize}
% \item Forward Injection (FI): Deep semantic features extracted from the RGB teacher’s backbone are injected into the prediction head of the thermal student, guiding it to learn geometric and topological priors that are richly encoded in RGB representations, as illustrated by the yellow dashed line in Figure \ref{pipeline_KD}.
% \item Reverse Injection (RI): Thermal features from the student are fed into the frozen RGB teacher’s prediction head. Leveraging the teacher’s stable, pretrained parameters, this reverse supervision enforces the student’s thermal features to be semantically interpretable within the teacher's learned representation space, as illustrated by the blue dashed line in Figure \ref{pipeline_KD}.
% \end{itemize}

\begin{itemize}
\item \textbf{Forward Injection (FI)}: The rich features from the RGB teacher's backbone are injected into the thermal student's prediction head. This path directly guides the student's decision-making process with high-quality, modality-rich features.
\item \textbf{Reverse Injection (RI)}: The learned features from the thermal student's backbone are injected into the frozen RGB teacher's prediction head. This path acts as a validation mechanism, forcing the student's features to be semantically compatible and interpretable by the stable, expert teacher.
\end{itemize}
These pathways generate two sets of cross-injected logits: $L_{ts}$, from the teacher's backbone and student's head (FI path), and $L_{st}$, from the student's backbone and teacher's head (RI path). The DIKD loss then aligns these cross-modal predictions with the original unimodal logits from the teacher ($L_{tt}$) and student ($L_{ss}$).

The Reverse Injection Loss validates the student's features:
\begin{equation}
\label{equation_CIKDT}
    L_{DIKD}^{ri} = -\frac{1}{N}\sum_{n}^{N}\sum_{k}^{K}\sum_{i}^{L} L_{ss}^i \log (L_{ts}^i).
\end{equation}
This loss compels the student's features ($F_{Thermal}$) to produce teacher-like predictions when processed by the teacher's head, thereby ensuring feature-level semantic alignment.

The Forward Injection Loss guides the student's head:
\begin{equation}
\label{equation_CIKDS}
    L_{DIKD}^{fi} = -\frac{1}{N}\sum_{n}^{N}\sum_{k}^{K}\sum_{i}^{L} L_{tt}^i \log (L_{st}^i).
\end{equation}
This loss encourages the student's head to behave like the teacher's head when given the same high-quality features from the teacher's backbone ($F_{RGB}$).

The total DIKD loss is a weighted sum:
\begin{equation}
L_{DIKD} = \lambda_{ri}  L_{DIKD}^{ri} + \lambda_{fi} L_{DIKD}^{fi},
\end{equation}
where $\lambda_{ri}$ and $\lambda_{fi}$  are hyper-parameters controlling the contribution of each directional loss. By enforcing this dual alignment, DIKD promotes robust, modality-invariant learning, significantly stabilizing and enhancing the effectiveness of the cross-modal distillation.
  
\subsection{The Model Compression Level}

% Upon obtaining a well-distilled thermal student model from the KTL, the framework proceeds to the second level named Model Compression Level (MCL), which focuses on reducing model complexity for real-time deployment or resource-constrained environments. In this stage, the thermal student model trained in KTL is reused as a new teacher, and its knowledge is distilled into a smaller thermal student model using only thermal inputs. A conventional knowledge distillation strategy is employed, combining both feature-level alignment and logits-level supervision to retain localization accuracy while significantly reducing model size and inference latency.

Following the KTL, the framework transitions to the Model Compression Level (MCL), which is dedicated to producing a highly efficient, deployment-ready model. In this stage, the high-fidelity thermal model produced by the KTL serves as a teacher, and its knowledge is distilled into a more compact student network. Unlike the KTL, which bridges the challenging RGB-to-thermal modality gap, the MCL operates entirely within the thermal domain. This intra-modal setting enables a more stable and focused learning process, aiming to improve model efficiency while preserving accuracy. To this end, we reuse the family of loss functions employed in KTL, specifically, the feature-based loss (Equation~\ref{equation_heatmap}), the logits-based loss (Equation~\ref{equation_logits_loss}), and the keypoint-based loss (Equation~\ref{equation_keypoints}), but adapt them to the thermal single-modality setting.
By leveraging the high-fidelity supervision of the thermal teacher, the MCL stage facilitates training of a compact, low-latency thermal model. This model is particularly suitable for real-world deployment in computationally constrained environments such as mobile or embedded systems for thermal face analysis.

\subsection{Time-Adaptive Distillation Decay Strategy}
The KTL and MCL loss functions are composite objectives that balance multiple tasks. Such multi-task learning, especially in the challenging cross-modal KTL stage, can lead to optimization conflicts and instability. To mitigate this, we adopt a time-adaptive distillation decay strategy, inspired by DWPose~\cite{yang2023effective}.

The core idea is to prioritize the distillation-related losses ($L_{fm}$, $L_{lg}$, $L_{DIKD}$) during the early stages of training, when the student benefits most from the teacher's guidance. As training progresses, the weight of these losses is gradually decayed, shifting the optimization focus towards the primary task loss, $L_{kp}$. This is controlled by a scaling factor $\gamma(t)$:
\begin{equation}
\label{equation_weight_decay}
\gamma(t) = 
\begin{cases}
1 - \dfrac{t - \alpha}{T}, & \text{if } t > \alpha \\
1, & \text{if } t \leq \alpha
\end{cases}
\end{equation}
where $t\in(1, \dots, T)$ is the current epoch, $T$ is the total number of epochs, and $\alpha$ is a warm-up period. The final loss functions are:
\begin{equation}
L_{KTL} = L_{kp} + \gamma(t)(\lambda_{fm} L_{fm} + \lambda_{lg} L_{lg} +\lambda_{dikd} L_{DIKD}),
\end{equation}
\begin{equation}
L_{MCL} = L_{kp} + \gamma(t)(\lambda_{fm} L_{fm} + \lambda_{lg} L_{lg}),
\end{equation}
where each $\lambda$ denotes a hyper-parameter controlling the relative importance of the corresponding loss term.
This adaptive schedule orchestrates a smooth transition from generalized knowledge absorption to task-specific fine-tuning, leading to more stable training and improved final performance.

\section{EXPERIMENTS}
\label{sec:exp}

\subsection{Datasets and Preprocessing}
Our experimental design leverages a combination of large-scale public datasets for robust knowledge transfer and multi-faceted evaluation.

\textbf{RGB Pretraining Datasets:} To equip our RGB teacher network with a strong understanding of facial geometry, we pretrain it on a composite dataset formed from two benchmarks. We utilize the facial keypoint subset of COCO-WholeBody \cite{jin2020whole}, a large-scale dataset providing diverse human poses, to learn a generalized facial structure. This is augmented with 300W \cite{sagonas2016300}, a classic in-the-wild facial alignment dataset, which helps the model adapt to challenging real-world variations in lighting, expression, and occlusion.

\textbf{Cross-Modal Training Dataset:} The cornerstone of our KTL is the T-FAKE dataset \cite{flotho2025t}. As the first large-scale synthetic dataset offering paired RGB and thermal facial images, it is uniquely suited for our distillation task. Its 200,000 image pairs, generated from over 2,000 subjects with balanced demographics and simulated environmental conditions, provide the rich, paired data necessary to learn robust cross-modal feature mappings.

\textbf{Evaluation Datasets:} We conduct a thorough evaluation of our final thermal models on two distinct datasets: the test set of T-FAKE and the real-world CHARLOTTE-ThermalFace dataset \cite{ashrafi2022charlotte}. 
To facilitate a fine-grained analysis of model performance under different conditions, we adopt the evaluation protocol from \cite{wood2021fake}, splitting the CHARLOTTE-ThermalFace dataset based on two primary criteria: image size and facial perspective. For size-based analysis, the dataset is partitioned into High, Low, and Mean subsets, corresponding to images with widths greater than, less than, and equal to the median of 200 pixels, respectively. For perspective-based analysis, it is divided into a Side subset, comprising profile views annotated with 43 landmarks, and a Front subset, containing all frontal views with 73 landmarks. In addition to these subsets, we also report performance on the Full dataset, which encompasses all images.

\textbf{Qualitative Real-World Assessment:} To evaluate our model's robustness beyond static benchmarks, we collected a supplementary in-the-wild thermal video dataset using a FLIR AX70 camera. This dataset features multiple subjects exhibiting a range of challenging, unscripted facial dynamics, including rapid blinking, prolonged eye closure, significant head pose variations, and diverse facial expressions. This allows for a qualitative assessment of our model's stability and performance under realistic conditions.

\textbf{Keypoint Format Unification:} Since our evaluation datasets use different annotation conventions, we standardize them for consistent analysis. We map the landmark definitions from CHARLOTTE-ThermalFace (73/43 points) to our unified 68-point schema, which is consistent with the COCO and 300W standards. During evaluation on specific subsets like Side, we use the available corresponding landmarks.

\subsection{Evaluation Metrics}
To provide a holistic assessment of our proposed framework, we employ a comprehensive set of metrics that evaluate three key aspects of performance: landmark detection accuracy, model complexity and efficiency, and training process stability.

\textbf{Landmark Detection Accuracy:} We evaluate the performance of facial landmark detection models using the Normalized Mean Error (NME)~\cite{zhu2012face, wood2021fake, sagonas2016300}, which measures the average Euclidean distance between predicted and ground-truth keypoints, normalized by a reference facial distance:
\begin{equation}
\label{NME}
\mathrm{NME} = \frac{1}{K} \sum_{i=1}^{K} \frac{\left| \hat{p}_i - p_i \right|_2}{d},
\end{equation}
where $K$ is the number of facial landmarks, $\hat{p}_i$ and $p_i$ denote the predicted and ground-truth coordinates of the $i$-th keypoint, and $d$ is a normalization factor that compensates for scale variations across samples.
For the T-FAKE and SF-TL54 datasets, we adopt the inter-ocular distance as the normalization term, denoted as NME-IO, following prior works~\cite{sagonas2016300, wood2021fake}.
However, the CHARLOTTE-ThermalFace dataset frequently includes only one visible eye in thermal frames. In this case, we follow the convention of~\cite{zhu2012face} and use the diagonal of the ground-truth bounding box as the normalization factor. We report this variant as NME-W/H.

\textbf{Model Complexity and Efficiency:} To evaluate the computational requirements and suitability of our models for deployment, we report the total number of trainable Parameters (Params), which serves as a proxy for static memory footprint, and the Giga Floating-Point Operations (GFLOPs), which quantifies the theoretical computational cost of a single forward pass. Together, these metrics reflect the model's static and dynamic complexity, respectively.

\textbf{Training Stability Analysis:} To quantitatively validate our claim that Reverse Injection (RI) stabilizes training, we analyze the L2 norm of the gradients during the initial optimization phase. Let $g_t = | \nabla L_t |_2$ be the gradient norm at training step $t$. We assess the stability of the sequence ${g_t}$ using the following statistical metrics:
\begin{itemize}
    \item \textbf{Standard Deviation (Std Dev)}: Measures the absolute dispersion or volatility of the gradient norms around their mean. A lower value indicates more consistent gradient magnitudes.
% \begin{equation}
% \sigma_g = \sqrt{\frac{1}{N} \sum_{t=1}^{N} (g_t - \bar{g})^2}
% \end{equation}

    \item \textbf{Coefficient of Variation (CV)}: 
A normalized measure of dispersion, defined as the ratio of the standard deviation to the mean, expressed as a percentage. It allows for comparing volatility across different scales.
% \begin{equation}
% \mathrm{CV} = \left( \frac{\sigma_g}{\bar{g}} \right) \times 100%
% \end{equation}

    \item \textbf{Mean Step Change (MSC)}: Quantifies the average absolute change between consecutive gradient norms. A lower value indicates a smoother, less oscillatory training process.
% \begin{equation}
% \mathrm{MSC} = \frac{1}{N-1} \sum_{t=2}^{N} |g_t - g_{t-1}|
% \end{equation}
    \item \textbf{Coefficient of Determination ($R^2$)}: Measures how well the gradient norm sequence fits a linear trendline. A higher $R^2$ value suggests a more predictable and less chaotic training dynamic.
% \begin{equation}
% R^2 = 1 - \frac{\sum_{t=1}^{N} (g_t - f_t)^2}{\sum_{t=1}^{N} (g_t - \bar{g})^2}
% \end{equation}
% where $f_t$ is the value predicted by the linear regression model at step $t$.
\end{itemize}
Together, these metrics provide a robust quantitative framework to evaluate not only the final accuracy and efficiency of our models but also the quality and stability of the optimization process itself.

\subsection{Implementation Details}
Our framework is implemented in PyTorch, building upon the open-source MMPose \cite{mmpose2020} repository. We use the RTMPose \cite{jiang2023rtmpose} family of models as the backbone architecture for both teacher and student networks, allowing for easy scaling of model size. All experiments were conducted on an NVIDIA RTX 4090 GPU.

\textbf{RGB Teacher Pretraining:}  To build a powerful and knowledgeable teacher, we first pretrain a suite of RTMPose models of varying sizes (RTMPose-m, RTMPose-l, RTMPose-x) on a composite RGB dataset combining COCO-WholeBody \cite{jin2020whole} and 300W \cite{sagonas2016300}. This pretraining equips our teacher models with a robust understanding of facial landmark configurations and their appearance variations in unconstrained environments, making them highly effective knowledge sources for the subsequent distillation stages.

\textbf{Knowledge Transfer Level (KTL):} In this level, a pretrained RGB teacher model (e.g., RTMPose-l) is frozen and used to supervise a thermal student of the same or smaller size. The student is trained using the paired RGB-thermal images from the T-FAKE training set. The complete KTL loss function, which includes our DIKD mechanism, is employed. The loss weights are carefully balanced, with the keypoint supervision weight set to $\lambda_{kp}=0.1$ and all distillation-related weights (for $L_{fm}, L_{lg}, L_{fi}, L_{ri}$) set to $10^{-3}$.

\textbf{Model Compression Level (MCL):} After the KTL, the resulting high-fidelity thermal model serves as a new frozen teacher. A final, more lightweight thermal student (e.g., RTMPose-s or RTMPose-t) is then trained using only the thermal images from T-FAKE. For this intra-modal distillation, the loss weights are set to $\lambda_{kp}=0.1$ for direct supervision and $10^{-3}$ for the feature ($L_{fm}$) and logits ($L_{lg}$) distillation terms.

For both stages, we adopt the original RTMPose optimization settings (AdamW optimizer\cite{loshchilov2017decoupled}, cosine annealing learning rate schedule) and train for 150 epochs with a batch size of 64. We apply early stopping based on validation set performance to reduce overfitting and accelerate training.

\subsection{Framework Validation}
To validate the overall effectiveness and practicality of our proposed MLCM-KD framework, we conduct a series of experiments, including quantitative, qualitative comparisons on benchmark datasets and evaluations of inference efficiency across different deployment scenarios.

\subsubsection{Benchmark Comparison}
To provide a rigorous and comprehensive evaluation of our proposed method, we conduct a benchmark comparison against a wide array of baseline models. To ensure a fair assessment, all models, including our own, are trained and evaluated under identical conditions within the unified MMPose framework. Our comparison is structured into three main categories:

\textbf{Supervised-Only Baselines:} To establish a crucial performance baseline using only direct supervision on thermal data, we benchmark against an extensive and diverse set of models. Our selection is structured into three distinct methodological paradigms to cover the landscape of modern facial landmark detection. First, we evaluate the classic top-down regression paradigm by testing a ResNet-50\cite{he2016deep} backbone, both with a standard loss and with specialized losses like WingLoss\cite{feng2018wing} and SoftWingLoss \cite{lin2021structure}. Second, we include a range of well-established heatmap-based architectures that were designed specifically for facial landmark or general keypoint estimation, such as SCNet-50\cite{liu2020improving}, HourglassNet \cite{newell2016stacked}, HRNetv2\cite{sun2019deep, zhang2020distribution}, and the efficient MobileNetV2\cite{sandler2018mobilenetv2}. Finally, recognizing that many state-of-the-art architectures originate from whole-body pose estimation, we adapt several powerful models for the facial landmark task. This group includes modern efficient designs like CSPNeXt\cite{chen2024cspnext} and S-ViPNAS \cite{xu2021vipnas}, as well as the full, scalable family of RTMPose variants\cite{jiang2023rtmpose}, which serves as the foundation for our own method.

\textbf{Conventional Cross-Modal Knowledge Distillation Baselines:}  To fairly isolate the benefits of our novel MLCM-KD mechanism, we establish a strong set of conventional Cross-Modal knowledge distillation baselines. These models are trained using only standard distillation losses ($L_{fm}$ and $L_{lg}$) from a pretrained RGB teacher. We denote these models with a TeacherStudent suffix (e.g., RTMPose-xl), signifying that a powerful RTMPose-x teacher distills its knowledge into a smaller RTMPose-l student. This allows for a direct comparison against traditional KD.

\textbf{MLCM-KD (Ours)}: Finally, our proposed MLCM-KD models represent the full application of our two-level framework. These models also follow the Ours-TeacherStudent naming convention (e.g., Ours-xl). This category is intended to showcase the full potential and synergistic benefits of our decoupled design and bidirectional knowledge distillation strategy.

\subsubsection{Inference Speed Test}
To assess the practical deployability and computational efficiency of our models, we conduct a comprehensive inference speed benchmark. Following the MCL stage, each lightweight student model is exported to the ONNX format and subsequently optimized using NVIDIA's TensorRT for peak performance. We measure the average latency (in milliseconds) and throughput (in frames-per-second, FPS) under a simulated real-time scenario: processing a single image at a resolution of $256\times256$ with a batch size of 1. The benchmark is executed across a diverse array of hardware platforms to provide a view of performance, spanning from high-end servers to resource-constrained edge devices:
\begin{itemize}
    \item \textbf{High-end GPU}:  An NVIDIA GTX 4090, representing a server-side or high-performance computing environment.
    \item \textbf{Low-end GPU}: An NVIDIA GeForce GTX 1660 Ti, representing a typical consumer-grade desktop setup.
    \item \textbf{High-Performance CPU}: An Intel Core I9-14900KF, simulating deployment on a modern, high-end desktop without dedicated GPU acceleration.
    \item \textbf{Standard CPU}: An Intel Core I7-11700, representing a typical laptop or older desktop, to evaluate performance on more common, less powerful processors.
    \item \textbf{Mobile/Edge Device}: An NVIDIA Jetson AGX Orin (64GB RAM), an ARM-based embedded platform representing a typical deployment scenario for high-end, resource-constrained systems.
\end{itemize}
This comprehensive evaluation confirms that our proposed framework not only achieves state-of-the-art accuracy but also enables efficient, real-time deployment under a variety of practical, resource-constrained conditions.

\subsection{Ablation Study}
To validate the effectiveness of our proposed framework and dissect the contributions of its core components, we conduct a series of ablation studies on the T-FAKE dataset. Our investigation is structured to answer two fundamental questions: (1) How does our full knowledge transfer strategy compare against conventional distillation methods? and (2) What are the specific roles and synergistic effects of the forward and reverse pathways within our novel DIKD mechanism?

\subsubsection{Knowledge Transfer Strategies}
First, we establish the overall superiority of our approach by comparing it against standard distillation techniques. To ensure a fair comparison, all variants in this study utilize the same teacher-student architecture and optimization settings. We evaluate the following strategies:
\begin{itemize}
    \item \textbf{Feature Distillation (FD) Only}: A baseline that aligns only the intermediate feature representations between teacher and student backbones using the $L_{fm}$ loss.
    
    \item \textbf{Logits Distillation (LD) Only}: A baseline that transfers knowledge only at the prediction level by matching the logits distributions using the $L_{lg}$ loss.
    
    \item \textbf{FD + LD}: A strong baseline that combines both feature and logits distillation, representing a typical transfer approach.
    
    \item \textbf{Ours}: Our full KTL, which augments the Conventional KD baseline with our proposed DIKD mechanism, to demonstrate the performance uplift gained from bidirectional supervision.
\end{itemize}

\subsubsection{Dissecting the DIKD Mechanism}
Having established the overall benefit of our approach, we then perform a fine-grained analysis to isolate the individual contributions and interplay of the forward and reverse injection pathways. Starting from a baseline that includes all standard losses ($L_{kp}$, $L_{fm}$, $L_{lg}$), we incrementally add the components of our DIKD mechanism:
\begin{itemize}
    \item \textbf{Baseline (Conventional KD)}: Includes $L_{\text{fm}}$, $L_{\text{lg}}$, and $L_{\text{kp}}$ without any cross-modal structural coupling.
    
    \item \textbf{Baseline + Forward Injection (FI)}: We add only the $L_{DIKD}^{fi}$ loss to the baseline. This experiment is designed to isolate the benefit of directly guiding the student's prediction head with the teacher's high-quality features.

    \item \textbf{Baseline + Reverse Injection (RI)}: We add only the $L_{DIKD}^{ri}$ loss to the baseline. This critical experiment isolates the benefit of validating the student's features using the frozen teacher head.
    
    \item \textbf{Ours}: The complete model incorporating both $L_{DIKD}^{fi}$ and $L_{DIKD}^{ri}$. This final configuration is designed to reveal the synergistic effect of combining the accuracy-driving guidance of FI with the stability-enhancing validation of RI.
\end{itemize}

This detailed ablation allows us to attribute performance gains directly to our novel contributions and understand the underlying mechanics of our framework.

\section{Results}
\label{sec:results}

\subsection{Benchmark Result}

\renewcommand{\arraystretch}{1.3}
\begin{table*}[!t]
\centering
\caption{Quantitative comparison of supervised learning and knowledge distillation (KD) based approaches for thermal facial landmark detection. Results are reported with input size $256\times256$, including the Params (M), GFLOPs, and NME. The evaluation is conducted on the T-FAKE dataset and the CHARLOTTE-ThermalFace dataset, with NME analyzed across different quality levels (High, Mean, Low) and pose variations (Front, Side).}
\label{tab:benchmark}
\resizebox{\textwidth}{!}{%
\begin{tabular}{cc|c|c|c|ccccccc}
\hline
\multicolumn{2}{c|}{\multirow{2}{*}{Method}}                                        & \multirow{2}{*}{InputSize} & \multirow{2}{*}{Params(M)} & \multirow{2}{*}{GFLOPs} & \multicolumn{7}{c}{NME (↓)}                                                                                                 \\
\multicolumn{2}{c|}{}                                                               &                            &                            &                         & T-FAKE          & High            & Mean            & Low             & Front           & Side            & Full            \\ \hline
\multicolumn{1}{c|}{\multirow{16}{*}{Supervised Learning}} & ResNet-50              & $256\times256$             & 7.32                       & 34.02                   & 0.0383          & 0.1778          & 0.1697          & 0.1972          & 0.1999          & 0.2010          & 0.1801          \\
\multicolumn{1}{c|}{}                                      & ResNet-50+SoftWingLoss & $256\times256$             & 7.32                       & 34.02                   & 0.0378          & 0.1754          & 0.1681          & 0.1963          & 0.1385          & 0.1991          & 0.1789          \\
\multicolumn{1}{c|}{}                                      & ResNet-50+WingLoss     & $256\times256$             & 7.32                       & 34.02                   & 0.0375          & 0.1748          & 0.1675          & 0.1958          & 0.1379          & 0.1985          & 0.1781          \\
\multicolumn{1}{c|}{}                                      & SCNet-50               & $256\times256$             & 34.02                      & 7.12                    & 0.0362          & 0.1572          & 0.1396          & 0.1954          & 0.1430          & 0.1765          & 0.1714          \\
\multicolumn{1}{c|}{}                                      & HourglassNet           & $256\times256$             & 94.86                      & 28.71                   & 0.0382          & 0.1491          & 0.1644          & 0.1953          & 0.1367          & 0.1749          & 0.1646          \\
\multicolumn{1}{c|}{}                                      & Mobilenetv2            & $256\times256$             & 9.58                       & 2.16                    & 0.0411          & 0.1560          & 0.1694          & 0.1953          & 0.1385          & 0.1817          & 0.1828          \\
\multicolumn{1}{c|}{}                                      & HRNetv2                & $256\times256$             & 9.66                       & 4.72                    & 0.0362          & 0.1513          & 0.1383          & 0.1974          & 0.1367          & 0.1818          & 0.1835          \\
\multicolumn{1}{c|}{}                                      & HRNetv2-darkness       & $256\times256$             & 9.66                       & 4.72                    & 0.0342          & 0.1521          & 0.1369          & 0.1952          & 0.1375          & 0.1779          & 0.1760          \\
\multicolumn{1}{c|}{}                                      & CSPNeXt-m              & $256\times256$             & 4.13                       & 17.54                   & 0.0358          & 0.1535          & 0.1415          & 0.1992          & 0.1319          & 0.1771          & 0.1692          \\
\multicolumn{1}{c|}{}                                      & CSPNeXt-l              & $256\times256$             & 7.16                       & 32.45                   & 0.0351          & 0.1519          & 0.1401          & 0.1980          & 0.1302          & 0.1769          & 0.1670          \\
\multicolumn{1}{c|}{}                                      & S-ViPNAS-Res50         & $256\times256$             & 2.01                       & 7.29                    & 0.0371          & 0.1552          & 0.1439          & 0.1961          & 0.1351          & 0.1793          & 0.1718          \\
\multicolumn{1}{c|}{}                                      & RTMPose-t              & $256\times256$             & 0.58                       & 4.33                    & 0.0406          & 0.1613          & 0.1467          & 0.1959          & 0.1318          & 0.1974          & 0.1728          \\
\multicolumn{1}{c|}{}                                      & RTMPose-s              & $256\times256$             & 1.03                       & 6.79                    & 0.0377          & 0.1637          & 0.1516          & 0.2627          & 0.1600          & 0.2070          & 0.1976          \\
\multicolumn{1}{c|}{}                                      & RTMPose-m              & $256\times256$             & 2.73                       & 15.54                   & 0.0345          & 0.1503          & 0.1529          & 0.2371          & 0.1799          & 0.2035          & 0.1821          \\
\multicolumn{1}{c|}{}                                      & RTMPose-l              & $256\times256$             & 5.75                       & 30.26                   & 0.0325          & 0.1534          & 0.1388          & 0.2281          & 0.1401          & 0.1971          & 0.1801          \\
\multicolumn{1}{c|}{}                                      & RTMPose-x              & $256\times256$             & 10.45                      & 52.54                   & 0.0316          & 0.1503          & 0.1370          & 0.2142          & 0.1302          & 0.1858          & 0.1731          \\ \hline
\multicolumn{1}{c|}{\multirow{12}{*}{KD based}}            & RTMPose-mt (Conv. KD)  & $256\times256$             & 0.58                       & 4.33                    & 0.0372          & 0.1571          & 0.1438          & 0.2241          & 0.1442          & 0.1940          & 0.1855          \\
\multicolumn{1}{c|}{}                                      & RTMPose-ms (Conv. KD)  & $256\times256$             & 1.03                       & 6.79                    & 0.0359          & 0.1545          & 0.1419          & 0.2199          & 0.1412          & 0.1915          & 0.1818          \\
\multicolumn{1}{c|}{}                                      & RTMPose-lt (Conv. KD)  & $256\times256$             & 0.58                       & 4.33                    & 0.0368          & 0.1561          & 0.1425          & 0.2215          & 0.1429          & 0.1923          & 0.1840          \\
\multicolumn{1}{c|}{}                                      & RTMPose-ls (Conv. KD)  & $256\times256$             & 1.03                       & 6.79                    & 0.0353          & 0.1539          & 0.1408          & 0.2162          & 0.1398          & 0.1891          & 0.1801          \\
\multicolumn{1}{c|}{}                                      & RTMPose-lm (Conv. KD)  & $256\times256$             & 2.73                       & 15.54                   & 0.0326          & 0.1515          & 0.1384          & 0.2035          & 0.1325          & 0.1818          & 0.1731          \\
\multicolumn{1}{c|}{}                                      & RTMPose-xl (Conv. KD)  & $256\times256$             & 5.75                       & 30.26                   & 0.0318          & 0.1509          & 0.1375          & 0.2011          & 0.1308          & 0.1802          & 0.1705          \\
\multicolumn{1}{c|}{}                                      & \textbf{Ours-mt}       & $256\times256$             & 0.58                       & 4.33                    & 0.0365          & 0.1559          & 0.1428          & 0.2209          & 0.1426          & 0.1927          & 0.1837          \\
\multicolumn{1}{c|}{}                                      & \textbf{Ours-ms}       & $256\times256$             & 1.03                       & 6.79                    & 0.0351          & 0.1531          & 0.1401          & 0.2158          & 0.1399          & 0.1890          & 0.1798          \\
\multicolumn{1}{c|}{}                                      & \textbf{Ours-ls}       & $256\times256$             & 1.03                       & 6.79                    & 0.0342          & 0.1521          & 0.1392          & 0.2115          & 0.1381          & 0.1865          & 0.1775          \\
\multicolumn{1}{c|}{}                                      & \textbf{Ours-lt}       & $256\times256$             & 0.58                       & 4.33                    & 0.0359          & 0.1548          & 0.1415          & 0.2173          & 0.1415          & 0.1901          & 0.1812          \\
\multicolumn{1}{c|}{}                                      & \textbf{Ours-lm}       & $256\times256$             & 2.73                       & 15.54                   & 0.0315          & 0.1498          & 0.1371          & 0.2005          & 0.1311          & 0.1798          & 0.1709          \\
\multicolumn{1}{c|}{}                                      & \textbf{Ours-xl}       & $256\times256$             & 5.75                       & 30.26                   & \textbf{0.0304} & \textbf{0.1485} & \textbf{0.1355} & \textbf{0.1951} & \textbf{0.1288} & \textbf{0.1765} & \textbf{0.1665} \\ \hline
\end{tabular}%
}
\end{table*}

\begin{figure*}[!t]
\centering
\includegraphics[width=\textwidth]{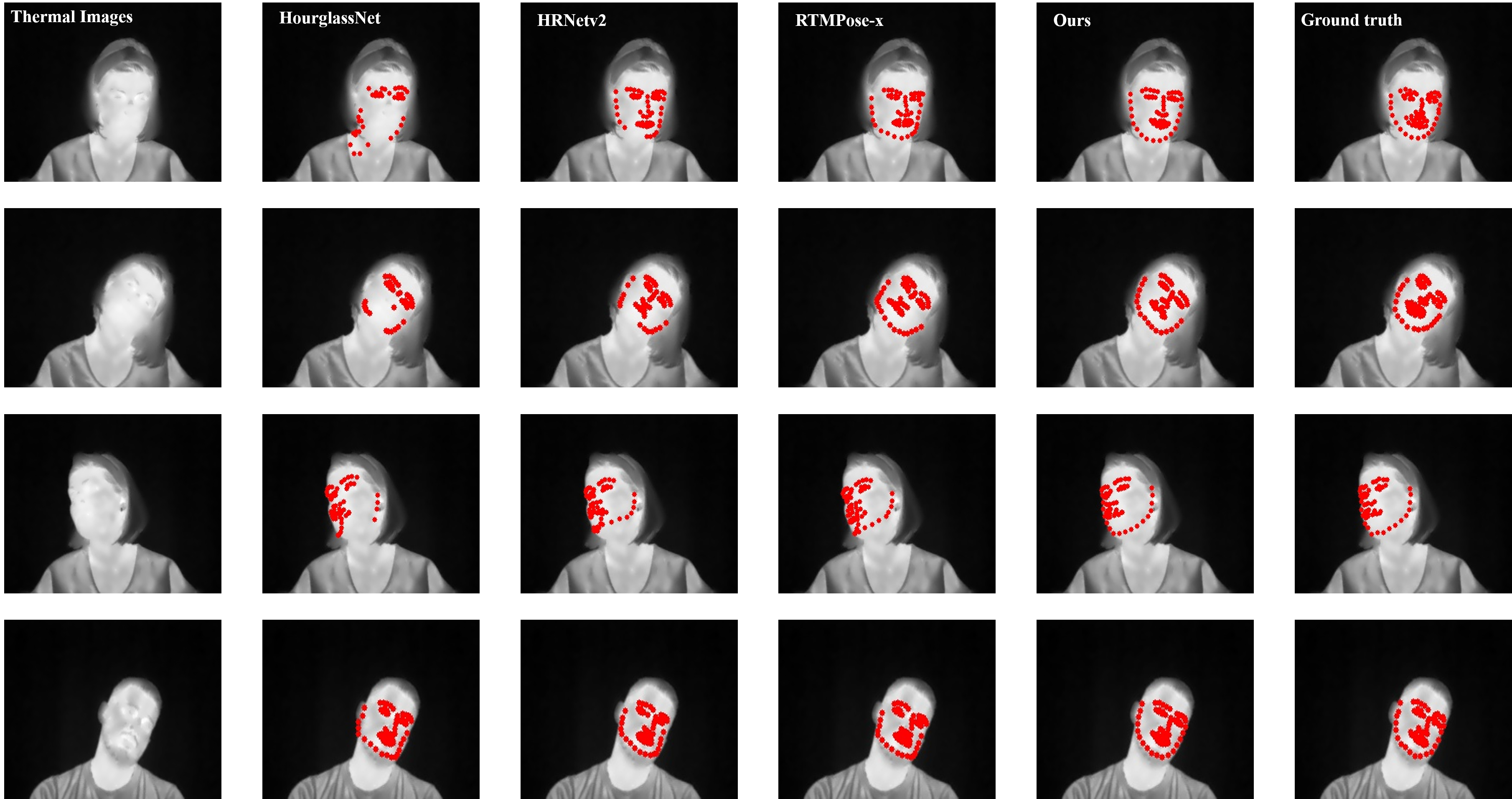}
\caption{Qualitative comparison of facial landmark detection results on thermal images. Each row shows different samples from the CHARLOTTE-ThermalFace dataset. From left to right: (a) input thermal image, (b) HourglassNet prediction, (c) HRNetV2 prediction, (d) RTMPose-x prediction, (e) our method, and (f) ground truth.}
% TODO : 
\label{fig:exp_menchmark}
\end{figure*}

\begin{figure}[!t]
\centering
\includegraphics[width=\linewidth]{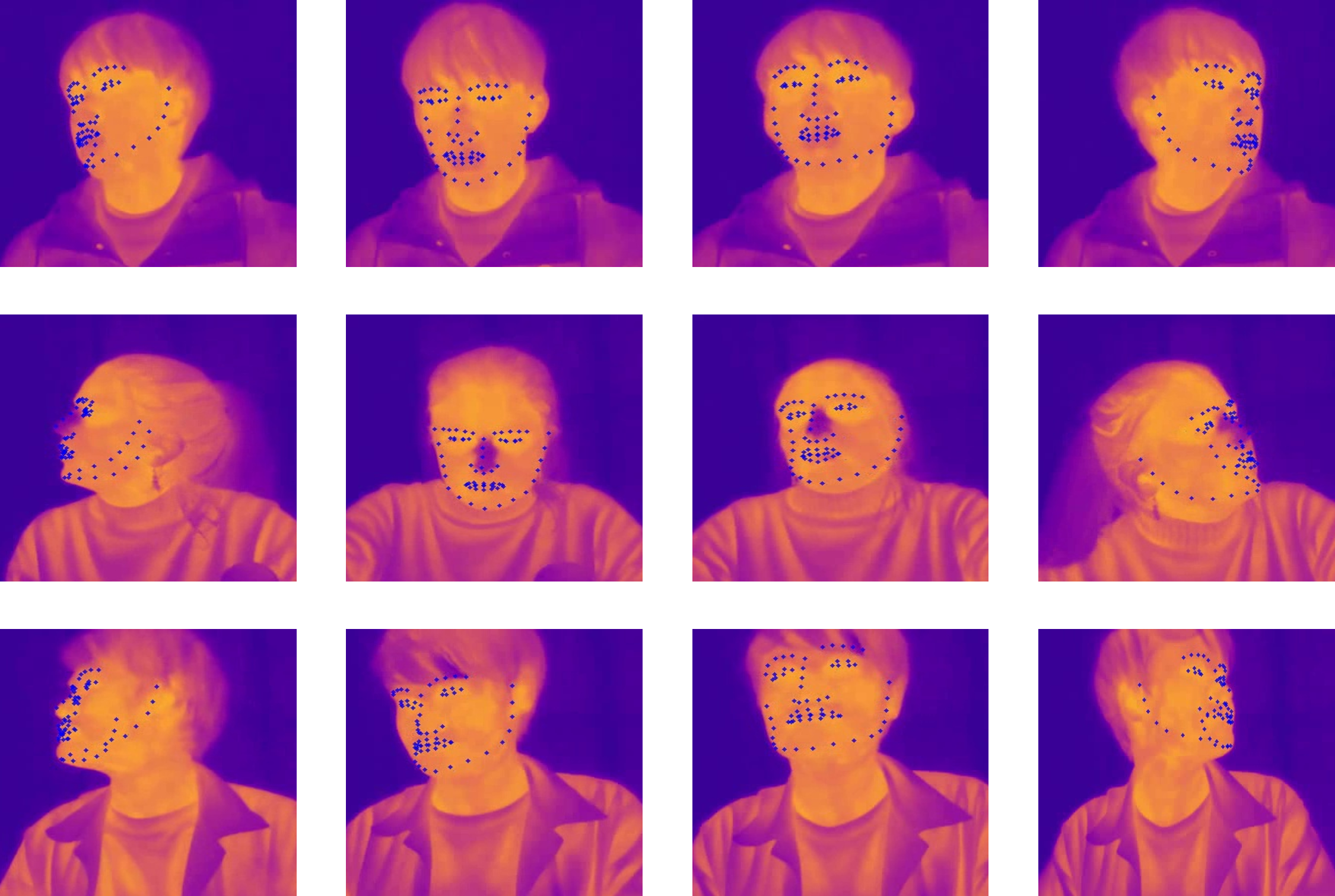}
\caption{Qualitative results of facial landmark detection on thermal images. Each row shows the same subject under different head poses, with facial landmarks estimated by our method. Samples are from our dataset captured with a FLIR AX70 camera.} %\textcolor{red}{TODO: Change the color of the facial landmarks in the thermal image to another color, such as cyan.}
\label{fig:exp_ours}
\end{figure}
 
This section presents a comprehensive quantitative and qualitative evaluation of our MLCM-KD framework against a wide array of state-of-the-art baselines. The primary results are summarized in Table \ref{tab:benchmark}, which details the performance across the T-FAKE and CHARLOTTE-ThermalFace datasets, alongside model complexity metrics.

Our first analysis focuses on models trained with direct supervision on thermal data, which establishes a performance baseline without cross-modal knowledge. As shown in Table \ref{tab:benchmark}, these models exhibit a wide range of performance, highlighting the inherent difficulty of the thermal FLD task. Heavyweight heatmap-based models like HourglassNet and SCNet-50 achieve competitive results (e.g., 0.1646 and 0.1714 ``Full'' NME respectively), benefiting from their powerful feature representation capabilities. However, their large model sizes make them less suitable for deployment. Moreover, lightweight models like RTMPose-s (0.1976 ``Full'' NME) and RTMPose-t (0.1728 ``Full'' NME) show a significant drop in accuracy, indicating that their limited capacity is insufficient to learn robust representations from low-contrast thermal data alone. Regression-based methods, such as ResNet-50+WingLoss, perform poorly (0.1781 ``Full'' NME) despite their high computational cost (34.02 GFLOPs), underscoring the general superiority of heatmap-based approaches for this task. These supervised results collectively demonstrate a clear accuracy-efficiency trade-off and set a challenging baseline for our knowledge distillation methods to surpass.

The introduction of knowledge distillation from a pretrained RGB teacher brings a dramatic improvement in performance, especially for lightweight models. The second block of Table \ref{tab:benchmark} shows the results for conventional, unidirectional KD. For instance, the RTMPose-s model, when trained with conventional KD (RTMPose-ls), sees its ``Full'' NME drop from 0.1976 to 0.1801. This confirms that even standard distillation can effectively transfer valuable priors from the RGB domain, helping smaller models achieve an accuracy that was previously only attainable by much larger supervised architectures. However, the performance gains are not uniform, and these models still struggle on challenging subsets, such as low-resolution images.

Our proposed MLCM-KD framework, powered by the DIKD mechanism, consistently sets a new state-of-the-art across nearly all model sizes and evaluation metrics. Ours-xl, achieves the best overall performance with a ``Full'' NME of 0.1665 on CHARLOTTE and 0.0304 on T-FAKE. This result is particularly remarkable as it outperforms even the heavyweight supervised models like HourglassNet (94.86M params) while using only 5.75M parameters.

To further investigate these performance differences, we provide a qualitative comparison in Figure \ref{fig:exp_menchmark}. We selected several representative baseline models, specifically HourglassNet, HRNetV2, and RTMPose-x, and visualized their predictions on challenging samples from the CHARLOTTE-ThermalFace dataset. The visualizations vividly corroborate our quantitative findings. As shown, baseline models frequently struggle with low-contrast and ambiguous thermal regions, often misplacing landmarks around the eyes, nose, mouth contours, and jawline. For example, HourglassNet tends to produce overly smooth and imprecise outlines, while even the powerful RTMPose-x can fail on profile views. In contrast, our method consistently generates more accurate and geometrically plausible landmark configurations that closely align with the ground truth. This visual evidence highlights our model's enhanced ability to interpret subtle thermal patterns by leveraging the rich, structural knowledge distilled from the RGB domain.

Finally, to validate the generalization capability of our framework beyond standard benchmarks, we evaluate our model on a custom, in-the-wild video dataset collected with a FLIR A70 thermal camera. As illustrated in Figure \ref{fig:exp_ours}, our method demonstrates remarkable robustness in these unconstrained, real-world scenarios. Even when faced with challenging head poses and subtle thermal gradients, our model maintains precise localization, particularly for geometrically complex areas like the eyes and the jawline, where other methods often fail. This strong performance on completely unseen data underscores the practical value and effective generalization of the knowledge transferred through our MLCM-KD pipeline.

\subsection{Inference Speed}
\begin{figure}[!t]
\centering
\includegraphics[width=\linewidth]{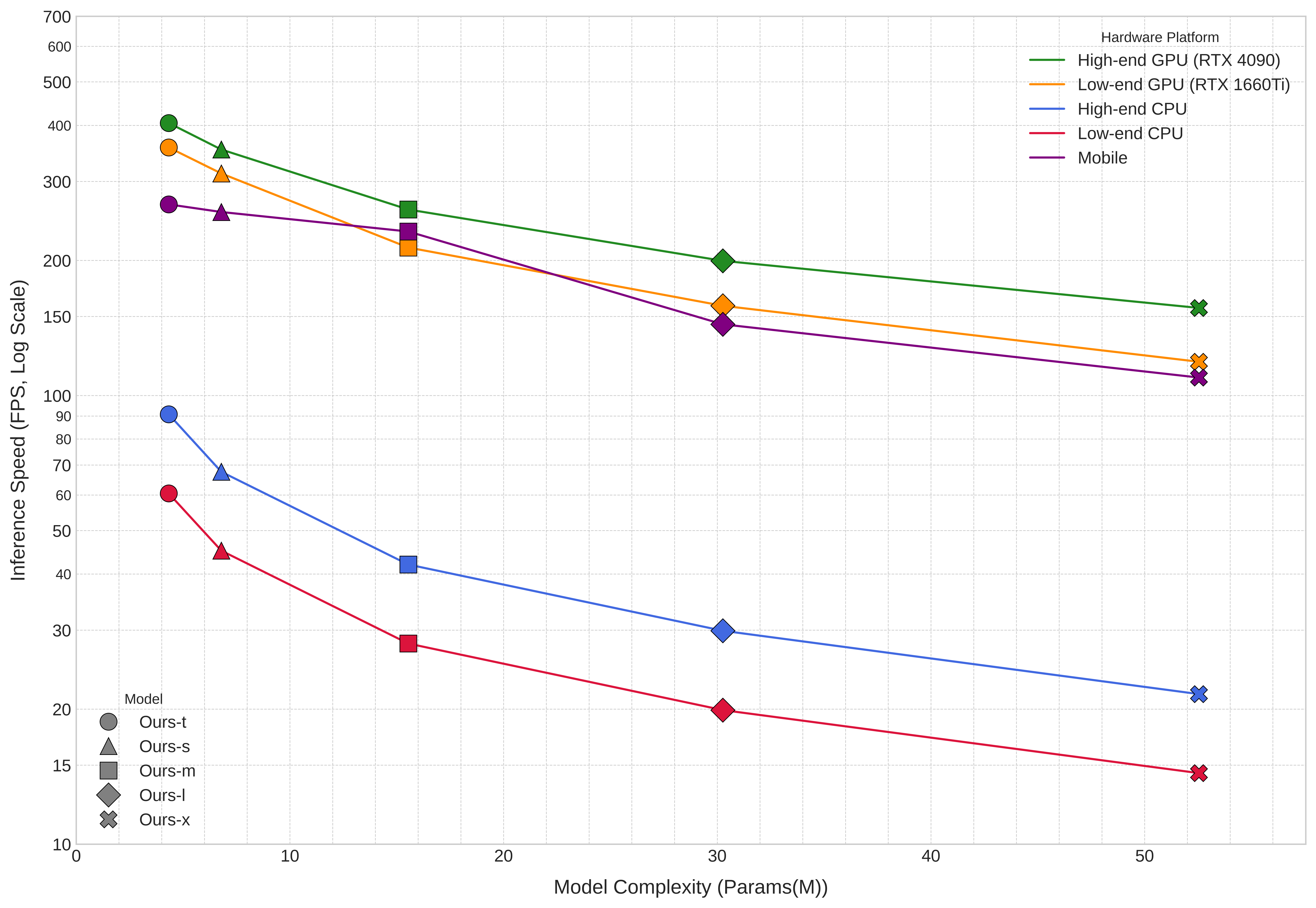}
\caption{Inference speed (FPS, logarithmic scale) versus model parameters for our model series across five hardware platforms: Mobile device, Standard CPU, High CPU, Low-end GPU (GTX 1660 Ti), and High-end GPU (GTX 4090).}
\label{fig:speed_complexity_tradeoff}
\end{figure}

A core objective of our framework is to produce models that are not only accurate but also computationally efficient for real-time deployment. We conduct a comprehensive inference speed benchmark across a spectrum of hardware, from high-end GPUs to resource-constrained edge devices, to validate this capability. The results, detailed in Table \ref{tab:speed} and visualized in Figure \ref{fig:speed_complexity_tradeoff}, robustly demonstrate the practical viability of our lightweight model series.

On modern GPU hardware, our models exhibit exceptional performance. Our most compact model, Ours-t, achieves an impressive 404.9 FPS on an NVIDIA GeForce RTX 4090 and maintains a highly practical 357.1 FPS on a consumer-grade NVIDIA GeForce GTX 1660 Ti. Even our largest model, Ours-x, sustains over 156 FPS on the high-end GPU and 118 FPS on the low-end GPU, confirming that the entire series is well-suited for GPU-accelerated pipelines.

Crucially, our framework's efficiency extends to CPU-only environments, which are common in real-world applications. On the high-performance Intel Core i9-14900KF, Ours-t and Ours-s deliver 90.8 FPS and 67.6 FPS respectively, comfortably surpassing the 60 FPS real-time threshold. Even on a standard processor like the Intel Core i7-11700, Ours-t and Ours-s remain real-time capable at 60.5 FPS and 45.1 FPS.

Performance on the edge AI platform is particularly noteworthy. Deployed on the NVIDIA Jetson AGX Orin, our models demonstrate remarkable efficiency. Ours-t reaches 266.7 FPS, and even our most complex model, Ours-x, runs at a remarkable 109.7 FPS. This outstanding performance on an ARM-based embedded system highlights the suitability of our models for demanding, on-device AI applications where both low latency and high accuracy are critical.
Figure \ref{fig:speed_complexity_tradeoff} illustrates the trade-off between model complexity and inference speed offered by our framework. The scalable performance across our model series allows practitioners to select an optimal configuration based on their specific accuracy needs and hardware constraints—from the ultra-fast Ours-t for mobile applications to the more powerful variants for server-side processing. This flexibility is a direct outcome of our decoupled distillation approach, which enables targeted and effective model compression.

In summary, our inference benchmarks validate that our framework successfully produces a family of models that bridge the gap between academic accuracy and real-world efficiency. We have demonstrated that our lightweight models not only achieve state-of-the-art precision but also meet the stringent latency demands for deployment on a wide array of computing platforms, from powerful servers to embedded edge devices.

\begin{table*}[!t]
\caption{Model complexity and inference speed benchmark on various hardware platforms.}
\centering
% \small
\renewcommand{\arraystretch}{1.3}
\resizebox{\textwidth}{!}{
\begin{tabular}{c|c|c|c|cc|cc|cc|cc|cc}
\hline
Method      & Input Size     & GFLOPs & Params(M) & \multicolumn{2}{c|}{Low CPU}                   & \multicolumn{2}{c|}{High CPU}                 & \multicolumn{2}{c|}{Low GPU (RTX 1660Ti)}       & \multicolumn{2}{c|}{High GPU (RTX 4090)}        & \multicolumn{2}{c}{Mobile}                      \\
            &                &        &           & (ms)                   & (fps)                 & (ms)                  & (fps)                 & (ms)                  & (fps)                   & (ms)                  & (fps)                   & (ms)                  & (fps)                   \\ \hline
RTMDet-nano & $320\times320$ & 0.31   & 0.99      & \multirow{2}{*}{16.53} & \multirow{2}{*}{60.5} & \multirow{2}{*}{11.0} & \multirow{2}{*}{90.8} & \multirow{2}{*}{2.80} & \multirow{2}{*}{357.14} & \multirow{2}{*}{2.47} & \multirow{2}{*}{404.86} & \multirow{2}{*}{3.75} & \multirow{2}{*}{266.67} \\
Ours-t      & $256\times256$ & 0.58   & 4.33      &                        &                       &                       &                       &                       &                         &                       &                         &                       &                         \\ \hline
RTMDet-nano & $320\times320$ & 0.31   & 0.99      & \multirow{2}{*}{22.18} & \multirow{2}{*}{45.1} & \multirow{2}{*}{14.8} & \multirow{2}{*}{67.6} & \multirow{2}{*}{3.20} & \multirow{2}{*}{312.50} & \multirow{2}{*}{2.83} & \multirow{2}{*}{353.36} & \multirow{2}{*}{3.89} & \multirow{2}{*}{256.43} \\
Ours-s      & $256\times256$ & 1.03   & 6.79      &                        &                       &                       &                       &                       &                         &                       &                         &                       &                         \\ \hline
RTMDet-nano & $320\times320$ & 0.31   & 0.99      & \multirow{2}{*}{35.76} & \multirow{2}{*}{28.0} & \multirow{2}{*}{23.8} & \multirow{2}{*}{42.0} & \multirow{2}{*}{4.68} & \multirow{2}{*}{213.68} & \multirow{2}{*}{3.85} & \multirow{2}{*}{259.74} & \multirow{2}{*}{4.31} & \multirow{2}{*}{232.02} \\
Ours-m      & $256\times256$ & 2.73   & 15.54     &                        &                       &                       &                       &                       &                         &                       &                         &                       &                         \\ \hline
RTMDet-nano & $320\times320$ & 0.31   & 0.99      & \multirow{2}{*}{50.37} & \multirow{2}{*}{19.9} & \multirow{2}{*}{33.6} & \multirow{2}{*}{29.9} & \multirow{2}{*}{6.31} & \multirow{2}{*}{158.48} & \multirow{2}{*}{5.01} & \multirow{2}{*}{199.60} & \multirow{2}{*}{6.94} & \multirow{2}{*}{144.09} \\
Ours-l      & $256\times256$ & 5.75   & 30.26     &                        &                       &                       &                       &                       &                         &                       &                         &                       &                         \\ \hline
RTMDet-nano & $320\times320$ & 0.31   & 0.99      & \multirow{2}{*}{69.61} & \multirow{2}{*}{14.4} & \multirow{2}{*}{46.4} & \multirow{2}{*}{21.6} & \multirow{2}{*}{8.41} & \multirow{2}{*}{118.90} & \multirow{2}{*}{6.38} & \multirow{2}{*}{156.73} & \multirow{2}{*}{9.12} & \multirow{2}{*}{109.65} \\
Ours-x      & $256\times256$ & 10.45  & 52.54     &                        &                       &                       &                       &                       &                         &                       &                         &                       &                         \\ \hline
\end{tabular}
\label{tab:speed}
}
\end{table*}

\subsection{Ablation Study}
\subsubsection{Knowledge Transfer Strategies}

\begin{table*}[ht]
\centering
\renewcommand{\arraystretch}{1.3}
\caption{Comparison of different distillation strategies using RTMPose-m/l/x architectures.}
\label{table_kd}
\begin{tabular}{c|c|c|ccccccc}
\hline
\multirow{2}{*}{Method}                 & \multirow{2}{*}{Teacher} & \multirow{2}{*}{Student} & \multicolumn{7}{c}{NME (↓)}                                                                                                 \\
                                        &                          &                          & T-FAKE          & High            & Mean            & Low             & Front           & Side            & Full            \\ \hline
Feature Distillation                    & RTMPose-m                & RTMPose-m                & 0.1353          & 0.1545          & 0.1486          & 0.2096          & 0.1486          & 0.1798          & 0.1748          \\
Logits Distillation                     & RTMPose-m                & RTMPose-m                & 0.0330          & 0.1583          & 0.1513          & 0.1862          & 0.1513          & 0.1567          & 0.1681          \\
FD + LD                                 & RTMPose-m                & RTMPose-m                & 0.0327          & 0.1532          & 0.1402          & 0.1959          & 0.1402          & \textbf{0.1520} & 0.1666          \\
\rowcolor{gray!15} \textbf{DIKD (Ours)} & RTMPose-m                & RTMPose-m                & \textbf{0.0325} & \textbf{0.1516} & \textbf{0.1356} & \textbf{0.1665} & \textbf{0.1356} & 0.1533          & \textbf{0.1594} \\ \hline
Feature Distillation                    & RTMPose-l                & RTMPose-l                & 0.1289          & 0.1493          & \textbf{0.1344} & 0.1992          & 0.1348          & 0.1593          & 0.1627          \\
Logits Distillation                     & RTMPose-l                & RTMPose-l                & 0.0347          & 0.1517          & 0.1409          & 0.1828          & 0.1261 & \textbf{0.1483} & 0.1601          \\
FD + LD                                 & RTMPose-l                & RTMPose-l                & 0.0341          & \textbf{0.1530} & 0.1437          & 0.1770          & 0.1283          & 0.1486          & 0.1629          \\
\rowcolor{gray!15} \textbf{DIKD (Ours)} & RTMPose-l                & RTMPose-l                & \textbf{0.0316} & 0.1533          & 0.1391          & \textbf{0.1725} & \textbf{0.1236} & 0.1487          & \textbf{0.1568} \\ \hline
Feature Distillation                    & RTMPose-x                & RTMPose-x                & 0.1278          & 0.1478          & 0.1325          & 0.1993          & 0.1321          & 0.1756          & 0.1656          \\
Logits Distillation                     & RTMPose-x                & RTMPose-x                & 0.0331          & 0.1473          & 0.1339          & 0.1683          & 0.1339          & 0.1479          & 0.1539          \\
FD + LD                                 & RTMPose-x                & RTMPose-x                & 0.0328          & 0.1476          & \textbf{0.1323} & 0.1703          & 0.1257          & 0.1462          & 0.1543          \\
\rowcolor{gray!15} \textbf{DIKD (Ours)} & RTMPose-x                & RTMPose-x                & \textbf{0.0301} & \textbf{0.1464} & 0.1324          & \textbf{0.1659} & \textbf{0.1251} & \textbf{0.1439} & \textbf{0.1536} \\ \hline
\end{tabular}  
\end{table*}

To isolate the sources of performance gains within our Knowledge Transfer Level, we conduct a detailed ablation study comparing various distillation strategies. The results, summarized in Table \ref{table_kd}, reveal a clear performance hierarchy. Feature Distillation alone yields the poorest results, particularly on the in-distribution T-FAKE test set. This indicates that simple feature mimicry, while a useful regularizer, is insufficient as a primary driver for cross-modal knowledge transfer in this task. In contrast, Logits Distillation provides a substantial performance boost. This is attributable to the strong spatial priors embedded in the SimCC formulation, where matching the teacher's logit distributions directly transfers precise localization knowledge. Combining FD and LD creates an even stronger baseline, confirming that supervising both the intermediate representation space and the final prediction space is complementary and effective.

Most importantly, the results demonstrate the superiority of our DIKD mechanism. Across all model scales, DIKD outperforms the strong `FD + LD' baseline, achieving the best overall performance  in nearly every configuration. For instance, with the RTMPose-m architecture, DIKD improves the `CHARLOTTE-ThermalFace Full' NME from 0.1666 to 0.1594. This consistent improvement validates our central hypothesis: that simply matching features and logits independently is not enough. The bidirectional structural alignment enforced by DIKD, in which the student's features are actively validated by the teacher's head and the student's head is guided by the teacher's features, establishes a more profound connection that bridges the modality gap more effectively.

\subsubsection{Impact of Distillation Loss Terms}
To precisely quantify the contribution of each component within our DIKD mechanism, we conduct a fine-grained ablation study, incrementally adding the forward ($L_{DIKD}^{fi}$) and reverse ($L_{DIKD}^{ri}$) injection losses to a strong baseline. The results, presented in Table \ref{tab:ablation}, reveal that the two pathways play  complementary roles: one primarily drives accuracy, while the other ensures training stability.

Our baseline, incorporating standard losses ($L_{fm}$, $L_{lg}$, $L_{kp}$), achieves a respectable NME of 0.0542. The introduction of FI ($L_{DIKD}^{fi}$) alone provides the single largest boost to accuracy, reducing the NME significantly to 0.0392. This confirms its role as the primary knowledge transfer channel. By injecting the teacher's rich, high-quality features into the student's prediction head, FI directly guides the student to learn a more accurate mapping from features to landmark coordinates. It effectively teaches the student what an ideal feature representation looks like for precise localization, helping it overcome the ambiguity of low-contrast thermal cues.

\begin{figure}[!t]
\centering
\includegraphics[width=\linewidth,height=5cm,keepaspectratio]{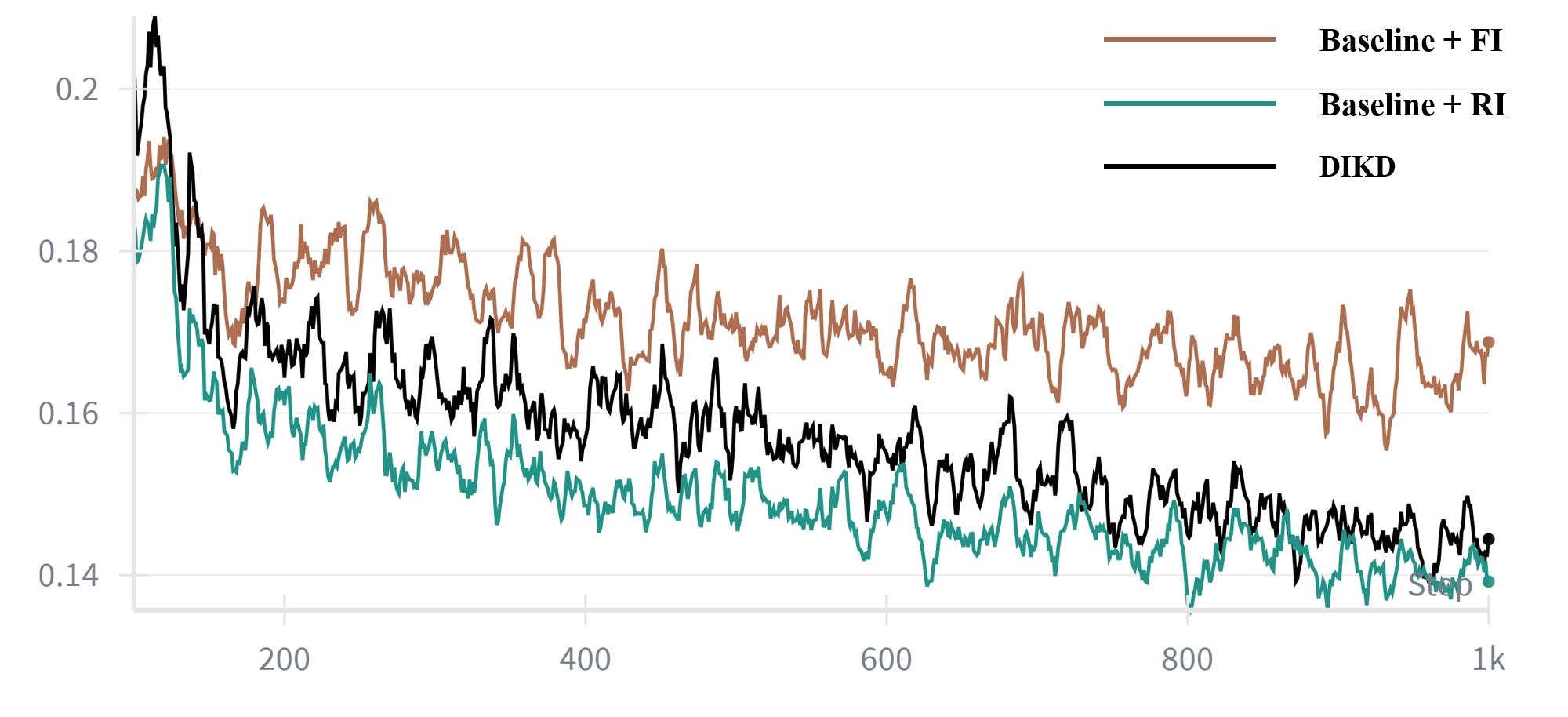}
\caption{Ablation study on loss terms and their impact on grad\_norm during training.}
\label{exp_KD}
\end{figure}

Conversely, adding RI ($L_{DIKD}^{ri}$) on its own yields a smaller but still notable accuracy improvement, reducing the NME to 0.0432. However, its primary contribution is not just accuracy but optimization stability. This is vividly illustrated in Figure ~\ref{exp_KD}, which plots the gradient norms during the crucial early stages of training. While the ``Baseline + FI'' models exhibit volatile gradient updates, the inclusion of RI leads to a visibly smoother and more stable gradient trajectory. We quantify this stabilizing effect in Table \ref{tab:stability_metrics}. The ``Baseline + RI'' configuration achieves the lowest standard deviation (0.0415) and coefficient of variation (22.93\%) in gradient norms, confirming it produces the most consistent training signal. This stability arises because RI forces the student's nascent thermal features to be interpretable by the frozen, expert teacher head, effectively anchoring the student's representation learning process and preventing erratic updates caused by noisy thermal inputs.

The effectiveness of our approach is most evident when both pathways are employed simultaneously. The full DIKD model, incorporating both $L_{DIKD}^{fi}$ and $L_{DIKD}^{ri}$, achieves the lowest NME of 0.0342. This result highlights a strong synergistic effect: the FI delivers rich geometric guidance for improved localization accuracy, while the reverse injection RI provides a semantic anchor that stabilizes and accelerates the learning process.

\renewcommand{\arraystretch}{1.8}
\begin{table}[!t]
\caption{Ablation study on loss terms and their impact on NME on the T-FAKE dataset.}
\centering
\begin{tabular}{c|c|c|c|c|c|c}
\hline
Method        & $L_{\text{fm}}$ & $L_{\text{lg}}$ & $L_{\text{kp}}$ & $L_{\text{DIKD}}^{fi}$ & $L_{\text{DIKD}}^{ri}$ & \textbf{NME (↓)} \\ \hline
Baseline      & \checkmark      & \checkmark      & \checkmark      & --                     & --                     & 0.0542           \\
Baseline + RI & \checkmark      & \checkmark      & \checkmark      & --                     & \checkmark             & 0.0432           \\
Baseline + FI & \checkmark      & \checkmark      & \checkmark      & \checkmark             & --                     & 0.0392           \\  \rowcolor{gray!15} 
Ours          & \checkmark      & \checkmark      & \checkmark      & \checkmark             & \checkmark             & \textbf{0.0342}  \\ \hline
\end{tabular}

\label{tab:ablation}
\end{table}

\begin{table}[htbp]
  \centering
  \caption{Quantitative Stability Metrics for Gradient Norms of Different Loss Components.}
  \label{tab:stability_metrics}
  \begin{tabular}{lcccc}
    \toprule
    Method & Std Dev & CV (\%) & Mean Step Change & $R^2$ \\
    \midrule
    $L_{\text{DIKD}}^{ri}$ & \textbf{0.0415} & \textbf{22.9306} & 0.0145 & \textbf{0.2247} \\
    $L_{\text{DIKD}}^{fi}$ & 0.0505 & 29.8237 & 0.0177 & 0.2959 \\
    \text{Ours}        & 0.0442 & 27.6011 & \textbf{0.0139} & 0.2784 \\
    \bottomrule
  \end{tabular}
\end{table}

\section{Conclusion}
\label{sec:conclusion}
In this paper, we presented MLCM-KD, a multi-level cross modal knowledge distillation framework that comprises the Knowledge Transfer Level (KTL) and the Model Compression Level (MCL) and is designed to overcome the modality gap between RGB and thermal imagery while meeting real-time deployment constraints. Our experiments first confirmed the limitations of existing approaches: heavyweight supervised models proved impractical for deployment, while lightweight models and conventional knowledge distillation techniques failed to achieve sufficient accuracy on challenging thermal data. Against this backdrop, our work has demonstrated a new state-of-the-art, establishing a robust and efficient solution to this problem.

% The success of our framework is rooted in our DIKD mechanism. We demonstrated that DIKD consistently outperforms standard feature and logit distillation, proving that a more profound structural alignment is necessary to bridge the modality gap. Furthermore, our analysis revealed the synergistic roles of its core components: experiments showed that forward injection is the primary driver of localization accuracy, while reverse injection provides a crucial stabilizing effect on the training dynamics. This synergistic design is the key to effectively transferring the rich geometric priors from the RGB teacher to the thermal student.

% Beyond setting a new benchmark for accuracy, we have proven the practical viability of our models. Our comprehensive inference speed tests show exceptional performance across a wide array of hardware, from achieving over 400 FPS on high-end GPUs to running in real-time on standard CPUs and, most notably, reaching over 266 FPS on the NVIDIA Jetson edge platform. This efficiency, combined with the model's demonstrated robustness on real-world data, confirms its suitability for real-world, resource-constrained applications.

The state-of-the-art accuracy of our framework is achieved through the KTL, which is powered by our novel DIKD mechanism. We demonstrated that DIKD consistently outperforms standard feature and logit distillation, proving that a more profound structural alignment is necessary to bridge the modality gap. Furthermore, our analysis revealed the synergistic roles of its core components: forward injection serves as the primary driver of localization accuracy, while reverse injection provides a crucial stabilizing effect on the training dynamics. This synergistic design is the key to effectively transferring the rich geometric priors from the RGB teacher.

Beyond the accuracy gains from the KTL, the MCL ensures the practical viability of our models. Our comprehensive inference speed tests show exceptional performance across a wide array of hardware, from achieving over 400 FPS on high-end GPUs to running in real-time on standard CPUs and, most notably, reaching over 266 FPS on the NVIDIA Jetson edge platform. This efficiency, combined with the model's demonstrated robustness on real-world data, confirms its suitability for real-world, resource-constrained applications.

MLCM-KD delivers more than a marginal performance gain; it establishes a principled and generalizable paradigm for transferring knowledge from data-rich to data-scarce modalities. Building on this foundation, our future work will extend the proposed framework to additional sensing domains, such as depth and near-infrared imaging, and adapt it to temporal video analysis, thereby enabling robust all-weather systems for facial tracking and behavior understanding. Moreover, we intend to broaden the scope of keypoint extraction from facial landmarks to whole-body skeletal landmarks, with the goal of supporting comprehensive multimodal analysis of human motion and social interactions.

\section*{Acknowledgments}
This work was supported by the Italian Workers’ Compensation Authority INAIL within the VIVA project.

\bibliographystyle{IEEEtran}
\bibliography{IEEERef}

\vfill

\end{document}